# A novel biomass fluidized bed gasification model coupled with machine learning and CFD simulation

Chun Wang[1]

**State Key Laboratory of Coal Combustion, Huazhong University of Science and Technology, 1037 Luoyu Road, Wuhan, Hubei, 430074, PR China**

E-mail: wangchunhust@163.com

Abstract: A coupling model of biomass fluidized bed gasification based on machine learning and computational fluid dynamics is proposed to improve the prediction accuracy and computational efficiency of complex thermochemical reaction process. By constructing a high-quality data set based on experimental data and high fidelity simulation results, the agent model used to describe the characteristics of reaction kinetics was trained and embedded into the computational fluid dynamics (CFD) framework to realize the real-time update of reaction rate and composition evolution.

*Keywords:* Biomass gasification, Machine learning (ML), Computational fluid dynamics (CFD), Kinetic parameters

1. **Introduction**

1. 1. **Background**

  Biomass thermochemical conversion technology refers to a series of processes in which biomass, such as crop straw, forestry waste and organic waste, are chemically converted into usable fuels or chemical raw materials under thermal conditions by some chemical reactions, such as pyrolysis and gasification[1,2]. Biomass thermochemical energy is one of important methods for the conversion of biomass energy, which has advantages such as low energy consumption, high efficiency of energy conversion, and so on, thus can be easily industrialized. Thermochemical conversion process is a complicated chemical reaction process. Efficiency, the content of effective components, and other performance parameters and features are all impacted by various factors, such as reactor types, biomass types and operation conditions[3-5]. In practical application process, biomass thermochemical techniques would meet with many challenges, such as feedstocks' diversity and changing of composition; product composition changes and difficulties in controlling quality. Therefore, exploring basic mechanism of biomass thermal conversion and guaranteeing high quality, long-term stable production of products' suitable control approach is very important.

# 1. 2. Problem Description

In order to improve the utilization of biomass, many research methods have been explored in the internal laws of thermochemical transformation of biomass[6,7]. Traditional experimental research requires a long process from raw material characterization, reactor adjustment, product detection to parameter optimization. Faced with challenges such as fluctuations in raw material components, complex reaction coupling, and difficulty in tar generation control, it usually takes months or even years to obtain local optimal operating conditions[8]. In recent years, in order to quickly screen process parameters in virtual space, numerical methods such as computational fluid dynamics (CFD), response surface method (RSM) and dynamic modeling have been introduced[9,10]. However, these methods still rely heavily on prior hypotheses and experimental calibrations, with limited ability to solve high-dimensional nonlinear and multi-scale coupling problems. In recent years, with the exponential accumulation of experimental data and the popularization of artificial intelligence, machine learning (ML) has provided a new paradigm for breaking the black box of "raw material processing products" with its automatic mining and prediction capabilities of high-dimensional feature spaces[11-13], which provides a powerful means for the study of thermochemical transformation of biomass.

Fig. 1 shows the amount of previously published works (only articles) by searching for "machine learning" and "biomass" on Web of science. It can be found from the figure that the literature on the research of machine learning in biomass thermochemical conversion shows three growth stages of "slow fast explosion" from 2010 to 2025: there are only sporadic explorations before 2015; From 2016 to 2019, with the popularity of in-depth learning, more than 100 studies on biomass thermochemical conversion were published annually using the in-depth learning framework; Since 2020, with the influence of physical information neural network, large-scale model, and the slogan of carbon neutrality in the policy, relevant research has entered a period of explosive growth, and has realized from simple auxiliary regression tools to intelligent optimization of the whole process. With the development of artificial intelligence, it can be predicted that there will be more kinds of machine learning applications in trying to explore the internal mechanism of biomass thermochemical conversion.

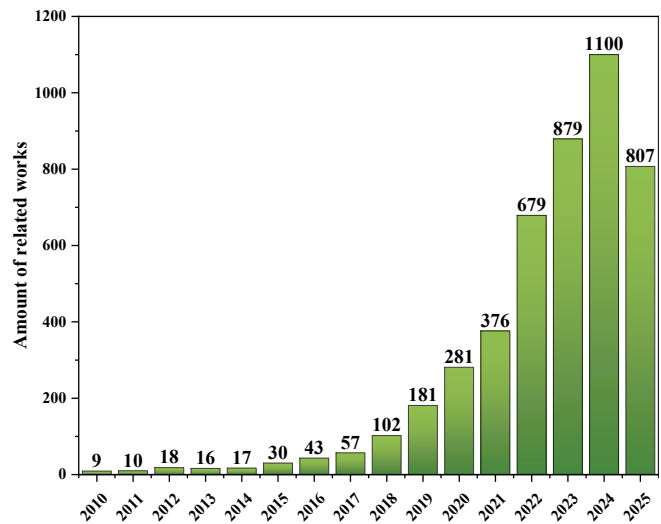

Fig. 1 Amount of related works published from 2010 to 2025.

## 2. Related Work

Machine learning has been widely used in biomass thermochemical processes to predict kinetic parameters, product yields and characteristics of biomass pyrolysis/gasification processes, and to deeply analyze the internal relationship between pyrolysis conditions and pyrolysis results. Based on 273 sets of experimental data, Cheenkahorn et al.[14] applied five machine learning models to predict and optimize the yield of biomass pyrolysis products (bio oil, biochar, and gas). The XGBoost model performed the best, with a training set $R^2$ exceeding 0.90. By analyzing the importance and sensitivity of features, the significant effects of biomass components (cellulose, hemicellulose, lignin) and pyrolysis parameters (temperature, heating rate) on product distribution were revealed. The accuracy and generalization ability of the model under different types of biomass were verified, providing a reliable basis for optimizing customized pyrolysis processes. Wang et al.[15] constructed and optimized four machine learning models to predict tar yield and hydrogen concentration during biomass gasification. The PSO-SVR model performed the best, with test set $R^2$ reaching 0.8701 and 0.9105, respectively. Through SHAP analysis and dual objective partial dependence diagram, it is revealed that temperature is a key factor affecting tar suppression and hydrogen enhancement, and the optimal gasification temperature is about 920-980 °C. In addition, measures such as using fluidized bed reactors, adding catalytic ore materials, using steam as gasification agent, and introducing catalysts can help achieve synergistic tar suppression and hydrogen production increase. Van et al.[16] proposed a physical information machine learning model that integrates carbon conservation constraints to predict the gas production composition and carbon flow direction of various biomass and plastic waste during the intermediate gasification process in fluidized beds. By comparing four data-driven models, XGBoost performed the best ($R^2$=0.99, RMSE=1.1), and based on this, a hybrid model was constructed by introducing physical constraints to significantly improve carbon closure while maintaining high prediction accuracy. SHAP analysis shows that the carbon content and feed rate of the feed are key characteristics that affect the output. Tasneem et al.[17] focused on modeling and predicting the solar driven biomass gasification (SDBG) process, and developed and compared various machine learning models, including random forest, RANSAC, SGD, ARD regression, and elastic network regression, to predict key parameters in continuous solar driven biomass gasification processes, such as $H_2$ and CO yields, carbon input and output rates, solar energy input and conversion efficiency, aiming to efficiently and cost effectively replace traditional experiments and mechanism analysis through machine learning methods.. All models were constructed in linear form, with elastic networks and ARD regression performing the best in terms of predictive formulas ($R^2$=0.999), while random forests and RANSAC performed the best in modeling accuracy ($R^2$=0.999).

The above studies focus on product prediction, without deep explanation of reaction process and reaction process, so many researchers combine machine learning with other research methods to deepen the understanding of biomass thermochemical reaction process. Weng et al.[18] developed a hard constrained physical information neural network (HC-PINN), through embedding the physical monotonicity knowledge into the model parameter by a constrained particle swarm optimization algorithm, in order to get high-precision, interpretable, generalizable predictions of the gas components from biomass gasification. As compared with traditional machine learning methods, HC-PINN not only avoids requiring a synthetic set to ensure the compliance of the physical laws but also ensures the entire model conforms to physical law under input space across all space throughout the training. Thus it can achieve very good performances with even insufficient samples or unknown samples. Cakar et al.[19] used Aspen Plus to simulate fixed bed and fluidized bed gasifiers, and systematically evaluated the synthesis gas composition and system effective performance of nine biomass and lignite mixed raw materials under different operating parameters (such as temperature, equivalence ratio, steam/biomass ratio, etc.). They combined machine learning methods (such as random forest, support vector regression, etc.) to establish a high-precision prediction model, achieving accurate estimation and optimization of gasification performance under small sample data, providing an efficient and data-driven new path for biomass co gasification clean hydrogen production. Zhong et al.[20] proposed a hybrid modeling method that combines computational fluid dynamics (CFD) and long short-term memory networks (LSTM) to predict the instantaneous mass flow rate and final yield of bio oil and other products during rapid pyrolysis of fluidized bed biomass. By training historical data generated by CFD, a reduced order model was constructed that includes LSTM layer, pooling layer, and fully connected layer, significantly reducing computation time by nearly 30% while maintaining high prediction accuracy. The research system optimized the key parameters of the LSTM model and extended it to predict other species such as gas, solid, and carbon at different pyrolysis temperatures, verifying the applicability and stability of the model. Kim et al.[21] proposed a data-driven method for rapid biomass pyrolysis prediction, which combines computational particle fluid dynamics (CPFD) simulation with machine learning regression models to predict the yield of pyrolysis products in a fluidized bed reactor. The study used reaction temperature and gas residence time as input features and compared eight models including linear regression, support vector regression, K-nearest neighbor, decision tree, random forest, and deep neural network. The results showed that traditional machine learning models (such as KNN, DT, RF) outperformed deep learning models in prediction accuracy and were highly consistent with CPFD results. This method not only improves modeling efficiency, but also provides new ideas for replacing traditional lumped reaction models and achieving accurate prediction of rapid pyrolysis processes.

It can be concluded from aforementioned studies that the most of the studies still focus on a certain part of the whole biomass thermochemical conversion process, only about the final product yield and ignore the intermediate reaction process, or only consider the partial reaction process in the intermediate reaction process, and do not combine machine learning with the overall reaction system to predict the reaction process in the biomass pyrolysis/gasification system more accurately.

## 3. Review of Representative literature 1

### 3.1. Summary of literature 1

Ji et al.[22,23] developed an autonomous framework for modeling biomass pyrolysis kinetics using Chemical Reaction Neural Networks (CRNN). By integrating fundamental physical laws (mass action, Arrhenius) into neural networks, CRNN simultaneously infers reaction pathways and kinetic parameters from Thermogravimetric Analysis (TGA) data without prior expert knowledge. Validated on cellulose pyrolysis, it achieved competitive accuracy with literature models while providing interpretable reaction mechanisms. This approach enables rapid, data-driven kinetic modeling for diverse solid fuels in fire and energy applications.

The proposed methodology leverages CRNN to autonomously model biomass pyrolysis kinetics by embedding fundamental physicochemical laws into a neural network architecture. The core principle lies in representing each chemical reaction as a neuron, where the inputs are the logarithmic concentrations of species (e.g., cellulose and intermediates) and the temperature. The weights encode the reaction orders (equal to stoichiometric coefficients) and Arrhenius parameters ($E_a$ and b). Biases represent the pre-exponential factors (A) on a logarithmic scale. The outputs are species production rates governed by the mass action law and Arrhenius equation. The CRNN stack forms a multi-reaction network (neurons are stacked into one hidden layer to formulate a CRNN for multi-step reactions) and enforces physical constraints, where cellulose (reactant) and volatiles (product) are specified to participate in unidirectional reactions. The reaction order is equal to the stoichiometric coefficient to ensure mass conservation. Regulating the catalytic/oxidative activity of oxygen through hyperparameters ($v'_{O_2}$).

The main implementation methods include:

(1) Data integration: Thermogravimetric analysis (TGA) residual mass data (normalized) was used as the training objective.
(2) Neural ODE coupling: A rigid ODE integrator (Rosenbrock23) was used to solve species concentration dynamics and compared the predicted residual mass with experimental data.
(3) Loss optimization: Stochastic gradient descent (Adam optimizer) is used to minimize the mean absolute error (MAE) between the predicted/measured quality, and gradient pruning is used to stabilize training.
(4) Sparsification: Hard threshold pruning (e.g., weights<0.185) eliminates redundant reactions and generates interpretable reaction networks.

## 3.2. Strong Points

This study made significant progress in chemical kinetics modeling by integrating fundamental physical and chemical laws into an autonomous framework of parameterized neural architecture. Its impact lies in derivation of interpretable kinetic models for complex solid fuels such as biomass. The novelty lies in directly embedding the law of mass action and Arrhenius equation into the network structure, enabling the simultaneous inference of reaction pathways, intermediate species, and kinetic parameters from thermogravimetric data. This method demonstrates depth by integrating a rigid ODE solver for reaction systems, optimizing physical constraints that ensure mass conservation, and sparsification for systems with clear mechanisms. In order to evaluate the rigor and accuracy of the proposed method, the researchers compared the results with another TGA dataset and existing cellulose models (such as the Broido-Shafizadeh scheme) using cross validation method under different pyrolysis/oxidation conditions. Using this method, the establishment time of reaction kinetic model can be shortened from several months to several hours, and the establishment of reaction kinetic model of fuel or industrial compound can be accelerated. This direct interpretation is applicable to the input combustion simulator. This study proposes a new method to link chemical reaction modeling with computer science through a physically guided data-driven paradigm, which shows the potential for change in combustion science, engineering and technology.

### 3.3. Weak Points

CRNN framework in this literature provides a new solution to kinetic modeling, but there are some problems as well. Firstly, assuming that the order is equal to the stoichiometric coefficient, while in reality, the reaction order is not necessarily integer. This oversimplification could make the model fail to capture the complex chemical behavior and deviation from stoichiometry of some reactions. For example, some reactions occur through non-ideal mechanisms, such as surface reaction and free radical intermediates, which makes the derivation of the formation and disappearance of the intermediate products somewhat speculative, without using experimental methods to confirm the existence of these products. Secondly, assuming ideal cases without considering heat and mass transfer. In this model, no restrictions were imposed on heat and mass transfer equations, so the model still has a gap with reality in some respects. Therefore, the next plan was further introduce physical law or coupled computational fluid dynamics (CFD) methods into the existing model framework in the future to investigate situations that may not meet the model assumptions. But that would dramatically increase the complexity of the model. Finally, grid search is an inefficient method for optimizing hyperparameters, taking 100 CPU hours to determine the number of species and reactions, pruning thresholds and others. Bayesian optimization or Evolutionary algorithm can be used instead of grid search to significantly reduce computational complexity, and an early stopping mechanism can be introduced to dynamically terminate some redundant searches based on validation set loss when the model starts to converge, thereby saving unnecessary computing power resources. Additionally, the model can learn knowledge from prior models by utilizing transfer learning by pre-training the CRNN using similar fuels (such as lignin), and later fine-tune and optimize parameters on the specific fuel.

### 3.4. Detailed Analysis

This manuscript provides a convincing framework for autonomous dynamics modeling, but lacks a clear rationale for the chosen hyperparameter search strategy. Section 3.3 describes grid search for species/response numbers and pruning thresholds, but does not compare this method with more effective methods such as Bayesian optimization or evolutionary algorithms, nor does it discuss computational trade-offs. This omission weakens the repeatability and scalability of the study. We suggest adding a section detailing the efficiency of hyperparameter optimization (such as computational cost and accuracy gain) and including a sensitivity analysis of pruning thresholds. Explicitly linking sparse thresholds to the stability of dynamic parameters (e.g., through confidence intervals).

## 4. Review of Representative literature 2

### 4.1. Summary of literature 2

Shi et al.[24] developed a novel hybrid framework that integrated machine learning models (LSSVM and ANN) with a chemical process platform (Aspen Plus) to address the limitations of modeling biomass gasification in a bubbling fluidized bed (BFB). The core principle replaces thermodynamic equilibrium assumptions and specific kinetics of raw materials with ML-driven rapid pyrolysis product prediction. Firstly, LSSVM and ANN models were trained on 197 gas and 310 carbon datasets to predict rapid pyrolysis products (carbon yield, gas composition) based on biomass properties and temperature, replacing thermodynamic equilibrium assumptions and feedstock-specific kinetics. Then, the products predicted by ML undergo rigorous mass and element balance calculations in the "pyrolysis product evolution aggregate balance" sub model, decomposing tar/charcoal into reactive components (such as C, $H_2$, CO, $CH_4$). Finally, reactive components were fed into Aspen's kinetically controlled gasification reactions to simulate the generation of synthesis gas.

The validation of six experimental datasets confirmed high accuracy and strong adaptability to different raw materials. Sensitivity analysis validated the predicted trends under different equivalence ratios, steam/biomass ratios, and temperatures. This method enables scalable high-fidelity modeling of biomass with unknown pyrolysis kinetics, and advances the design and optimization of gasifiers through innovative ML Aspen synergy.

### 4.2. Strong Points

This study has made significant progress in biomass gasification modeling through a novel hybrid ML Aspen framework, providing high scalability for various raw materials with unknown pyrolysis kinetics and achieving high validation accuracy (50% synthesis gas error ≤ ±10%). The depth of its technology lies in combining machine learning (LSSVM/ANN) with rigorous element balance sub-models and dynamic gasification simulations, rather than simply applying machine learning methods to predict product yields and achieve reliable trend predictions under different operating parameters. This framework provides substantial practical value for optimizing gasifier design.

### 4.3. Weak Points

This study exhibits several significant limitations that limit the robustness and industrial applicability of its methodology. Firstly, the modeling of tar is too simplistic, fixing naphthalene as the only inert tar component with a mass fraction of 20%, while failing to capture the dynamic cracking behavior of tar under different gasification conditions. This can lead to systematic errors in hydrogen and carbon balance prediction that worsen with changes in raw materials. Secondly, decoupling the carbon reactivity and gas-solid interaction from fluid dynamics and ignoring fluid dynamics conditions reduces the model fidelity of actual bubbling fluidized bed reactors, limiting the scalability of pilot scale systems. Thirdly, the black box nature of machine learning masks the interpretability of pyrolysis pathways, and the inconsistent characterization of biomass in literature datasets is not preprocessed, which reduces prediction accuracy and generalization ability. Overall, these gaps reduce prediction accuracy and limit the further application of the framework.

### 4.4. Detailed Analysis

Although the manuscript proposes a novel integration of machine learning and Aspen Plus modeling, the overall logical process needs improvement. The transition between the pyrolysis sub-model and the kinetic gasification module lacks a clear narrative bridge. A more structured explanation of how these components interact within the modeling framework would enhance readers' understanding.

Several technical terms and abbreviations have been introduced (such as "aggregate balance", "RYIELD", "SUNBRUNC"), but there is no clear definition, which may confuse readers who are not familiar with Aspen Plus or specific modeling terms. The first use of terminology or textual definitions will improve accessibility and the academic rigor of the manuscript.

Model assumptions and limitations, although the model has been validated based on experimental data, only brief discussions have been made on assumptions such as isothermal pyrolysis and fixed tar composition. A more critical reflection on the impact of these assumptions, especially under different operating conditions, would provide a more balanced perspective on the applicability and limitations of the model.

# 5. Proposed Method
## 5.1. Overall Design

In order to reveal the thermochemical conversion mechanism of biomass in thermochemical reactor more accurately and efficiently, shorten the modeling and solving time, and effectively guide the reactor design and scale-up process, a numerical simulation method based on the coupling of machine learning and computational fluid dynamics was proposed to simplify and efficiently solve the complex chemical reaction process. In view of the difficulties in obtaining the chemical reaction model, the weak generalization ability of the model and the strong dependence on specific working conditions in the traditional biomass thermochemical simulation process, a machine learning chemical reaction subject model based on high-quality experimental and simulation data was constructed. The model is trained based on high-precision data sets (including experimental data and high-precision CFD simulation results), which can effectively capture the characteristics of reaction kinetics and composition evolution, and has good prediction ability and extrapolation performance. In addition, the constructed machine learning agent model is closely coupled with the CFD framework. In the process of CFD calculation, the reaction rate and component change information predicted by the machine learning model are interpolated into the computational grid online through the coupling interface, replacing the traditional method of obtaining the reaction source term by solving complex partial differential equations. This method greatly reduces the time required for model debugging and parameter calibration, reduces the complexity of numerical solution and computational cost, and thus realizes the efficient and high-precision simulation of biomass thermochemical conversion process. The coupling strategy not only improves the simulation efficiency, but also provides reliable theoretical basis and technical support for reactor structure optimization and condition adjustment.

## 5.2. Dataset Collection
In view of the highly complex and multi-path parallel chemical reaction network involved in the biomass thermochemical conversion process, this study takes the biomass devolatilization process as a typical representative, and systematically describes the composition structure and data source of the constructed data set. The data set is composed of two parts, which aims to give consideration to the physical authenticity of the data and the universality of the working condition coverage, so as to provide high-quality and multi-dimensional input information for the training and verification of machine learning model.

The first part of the data comes from the published literatures, covering the component distribution of biomass pyrolysis products (such as CO, $CO_2$, $H_2$, $CH_4$, Tar, Char, etc.) and the thermogravimetric (TG) and differential thermogravimetric (DTG) curves obtained by traditional experimental means (such as thermogravimetric analysis, tubular furnace pyrolysis experiment, Fourier transform infrared spectroscopy, etc.). This kind of data not only reflects the characteristics of the final product of the devolatilization process, but also reveals the mass loss behavior and the evolution law of the reaction rate in the reaction process. It has a high degree of experimental repeatability and physical interpretability, and is the basic support for building a data-driven model. However, traditional experiments are difficult to cover boundary conditions such as high temperature, high pressure, rapid temperature rise or extreme reaction atmosphere. Therefore, the second part of the data introduces the virtual experiment results generated by high-precision CFD numerical simulation to fill the blank area of the experimental data. By coupling the detailed reaction kinetics mechanism with the multiphase flow model, the simulation results systematically scan the key operating parameters (such as heating rate, particle size, reactor wall heat flux, gas phase residence time, etc.) under controllable boundary conditions, so as to generate high-dimensional response data covering extreme conditions, which significantly expands the parameter space and generalization boundary of the data set.

In order to ensure the physical consistency and prediction reliability of machine learning model, the training set is composed of experimental data and simulation data to achieve data enhancement and space filling; The validation set only uses real experimental data to avoid the interference of simulation bias on model evaluation. Subsequently, the residual driven checking algorithm (such as physical constraint neural network, Bayesian error correction framework, etc.) can be introduced to dynamically correct the systematic error introduced by the simulation data, so as to improve the extrapolation ability and engineering applicability of the agent model. Finally, the input characteristics collected include key variables such as industrial analysis of biomass (moisture, volatile, fixed carbon and ash), element analysis (C, H, O, N, S), particle size distribution, temperature, system pressure, and gas phase component pressure. The output targets focus on the yield distribution of pyrolysis products (gas, liquid and solid three-phase products) and kinetic parameters (such as activation energy, pre-exponential factor, reaction order, etc.).

## 5.3. Model and Algorithm

Fig. 2 is the framework of machine learning coupled CFD modeling method based on Euler-Lagrangian framework, which is used to efficiently predict the reaction kinetic behavior in the process of biomass thermochemical conversion. By bidirectional coupling the discrete particle phase with the continuous gas phase, the framework extracts the flow field information in each grid and local reaction conditions such as particle temperature, particle size and heating rate in the flow field in real time, and inputs them as input variables into the pre trained machine learning model (such as random forest or neural network) to predict the corresponding reaction kinetic parameters (such as activation energy, pre exponential factor and reaction order). Subsequently, the predicted kinetic parameters are dynamically fed back to the CFD solver to update the devolatilization reaction rate and component evolution of the particle phase, realize the real-time correction of the reaction source term, and update the material concentration, pressure, temperature and other information in each grid calculation. Through two-way information interaction and iterative calculation, this process avoids the direct solution of complex partial differential equations involving chemical reactions in traditional methods, and significantly improves the simulation efficiency and prediction accuracy.

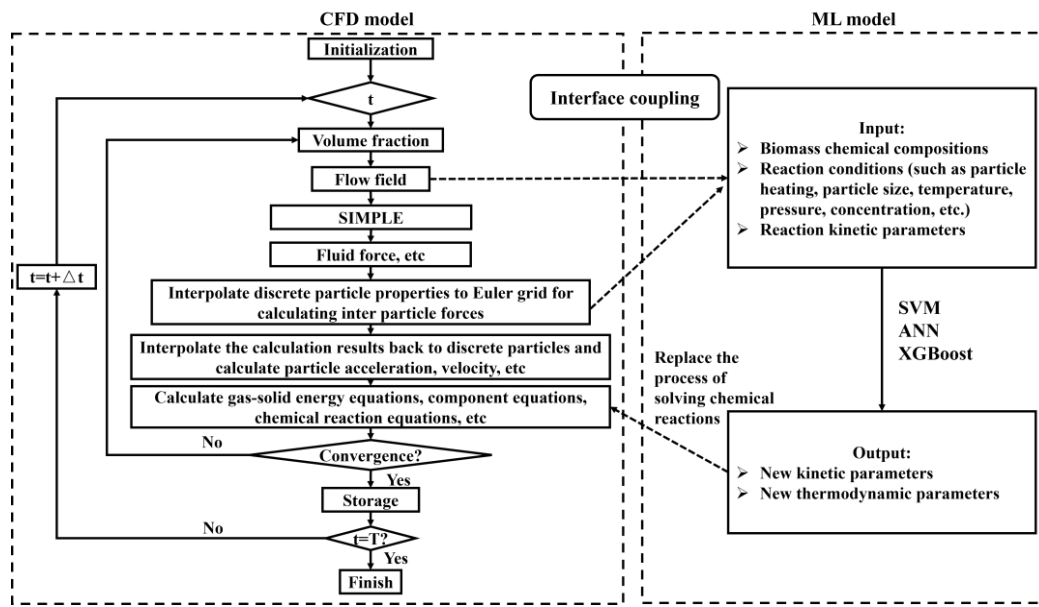

Fig. 2 Detailed program for simplifying chemical reaction calculations by coupling ML model with CFD simulation.

The main control equations for data interaction with the ML model in the calculation process of this method are as follows[25-27]:

Mass equation: $\dfrac{\partial(\theta_f \rho_f)}{\partial t} + \nabla(\theta_f \rho_f \mathbf{u}_f) = \delta \dot{m}_p$ (1)

Energy equation: $\dfrac{\partial(\theta_f \rho_f h_f)}{\partial t} + \nabla(\theta_f \rho_f \mathbf{u}_f h_f) = \theta_f(\dfrac{\partial p}{\partial t} + \mathbf{u}_f \cdot \nabla p) + \varphi - \nabla(\theta_f q) + \dot{q}_D + \dot{Q} + S_h$ (2)

Species conservation equation: $\dfrac{\partial(\theta_f \rho_f X_i)}{\partial t} + \nabla(\theta_f \rho_f \mathbf{u}_f X_i) = \nabla(\theta_f \rho_f D_i \nabla X_i) + \delta \dot{m}_i$ (3)

Particle energy conservation equation: $m_p C_p \dfrac{dT_p}{dt} = Q_{conv} + Q_{rad} - \Delta H_{rs}$ (4)

Reaction equations: $\sum\limits_{i=1}^{N} \gamma'_{1,r} M_i \Leftrightarrow \sum\limits_{i=1}^{N} \gamma''_{1,r} M_i$ (5)

$$r_i = k_{f,r} \prod_{i=1}^{N_s} C_i^{n_i} - k_{f,b} \prod_{i=1}^{N_s} C_i^{m_i}$$ (6)

$$\dfrac{dC_i}{dt} = \sum_{i=1}^{N_R} (\gamma''_{i,r} - \gamma'_{i,r}) r_i$$ (7)

In the equations, $\mathbf{u}_f$, $\rho_f$ and $\theta_f$ are the velocity vector, density and the volume fraction of gas in fluidized bed, respectively; $\delta \dot{m}_p$ is the source term of the mass production per unit volume from heterogeneous reactions; $p$ and $\tau$ are the gas pressure and stress tensor, respectively; $S_h$ is the conservative energy exchange between gas-solid phases; $\dot{Q}$ is the energy source per unit volume; $h_f$ is the gas enthalpy; $\varphi$ is the energy loss caused by viscous dissipation; $q$ and $\dot{q}_D$ are the gas heat flux and the enthalpy diffusion term; $D_i$ is the effective mass diffusion rate of species $i$; $X_i$ and $\delta \dot{m}_i$ are the mass fraction and mass transfer of species $i$ in chemical reactions; $C_p$ is the particle heat capacity; $Q_{conv}$, $Q_{rad}$ and $\Delta H_{rs}$ are the convective heat transfer, radiation heat transfer and chemical reaction heat; $k_{f,r}$ and $k_{f,b}$ are the forward and reverse reaction rate; $M_i$ is mole mass of component $i$; $C_i$ is the mole fraction of $i$; $N_R$ is the number of reactions participated by $i$. Among them, $\delta \dot{m}_p$, $X_i$, $\delta \dot{m}_i$, $\Delta H_{rs}$ and Reaction equations are the parameters that need to interact with the ML model.

The performances of the three proposed models for predicting kinetic parameters of biomass thermochemical conversion are evaluated with two statistical indicators, including the determination coefficient ($R^2$) and root mean square error (RMSE), which are defined as below[28,29]:

$$R^2 = 1 - \frac{\sum_{i=1}^{N}(y_{cal}^i - y_{exp}^i)^2}{\sum_{i=1}^{N}(y_{exp}^i - \overline{y}_{exp})^2} \tag{8}$$

$$RMSE = \sqrt{\frac{1}{N}\sum_{i=1}^{N}(y_{cal}^i - y_{exp}^i)^2} \tag{9}$$

where N is the number of the samples used in the training or test process, $y_{cal}^i$ is the predicted value of the $i^{th}$ sample, $y_{exp}^i$ is the experimental value of the $i^{th}$ sample, and $\overline{y}_{exp}$ is the average value of the entire training or testing samples.

## 6. Discussions and Future Work

By introducing a data-driven reaction dynamics surrogate model, the machine learning coupled CFD model framework proposed in this manuscript significantly reduces the dependence on traditional empirical parameters and the solution of complex differential equations, and effectively improves the computational efficiency and prediction accuracy. On the other hand, the data set based on experimental and high fidelity simulation data enhances the adaptability and generalization ability of the model under different working conditions, and has good engineering application potential. However, there are still shortcomings in this method. First, the prediction accuracy of the proxy model is highly dependent on the quality and coverage of the training data. If the training data are biased or not representative enough, it may lead to the instability of the model in practical application. In addition, the "black box" characteristics of the machine learning model weaken the interpretability of the model to a certain extent, which is not conducive to in-depth understanding of the reaction mechanism, and can only rely on the CFD module to supplement the analysis. Future research can be carried out in the following directions: first, expand the coverage of the data set, include more types of biomass and experimental and simulation data under extreme conditions, and improve the generalization ability of the model; The second is to extend the model to a complete gasification reaction network to realize the synchronous prediction of tar evolution, coke conversion and other key processes; The third is to introduce physical constraint mechanism or hybrid modeling strategy to enhance the interpretability and physical consistency of the model; The fourth is to improve the robustness and prediction reliability of model parameter identification by combining Bayesian optimization or uncertainty quantification methods. Through the above improvements, this method is expected to play a greater role in the intelligent simulation of biomass thermochemical conversion process and reactor optimization design.